\documentclass[10pt,twocolumn,letterpaper]{article}
\usepackage[T1]{fontenc}
\IfFileExists{main.bib}{}{}
\usepackage{times}
\usepackage{graphicx}
\usepackage{amsmath}
\usepackage{xcolor}
\usepackage{booktabs}
\usepackage{tabularx}
\newcommand{\papertablestyle}{\small\setlength{\tabcolsep}{3pt}}
\newcolumntype{Y}{>{\centering\arraybackslash}X}
\usepackage[numbers,sort&compress]{natbib}
\usepackage[hyphens]{url}
\usepackage[font=small,labelsep=period]{caption}
\usepackage{enumitem}
\usepackage[left=0.8125in,top=1in,textwidth=6.875in,textheight=8.875in,
  columnsep=0.3125in,headheight=0pt,headsep=0pt,footskip=30pt]{geometry}
\setlist[enumerate]{noitemsep,leftmargin=*,topsep=0pt}

\makeatletter
\renewcommand{\@maketitle}{%
  \newpage\null\vskip .25in
  \begin{center}
    {\Large\bfseries\@title\par}
    \vspace*{15pt}%
    {\large\lineskip .5em\begin{tabular}[t]{c}\@author\end{tabular}\par}
    \vskip .5em\vspace*{12pt}
  \end{center}%
}
\renewenvironment{abstract}
  {\centerline{\large\bfseries Abstract}\vspace*{12pt}\noindent\itshape\ignorespaces}
  {\vspace*{12pt}}
\renewcommand{\section}{\@startsection{section}{1}{\z@}%
  {-10pt plus -2pt minus -2pt}{7pt}{\large\bfseries}}
\renewcommand{\subsection}{\@startsection{subsection}{2}{\z@}%
  {-8pt plus -2pt minus -2pt}{5pt}{\fontsize{11}{12}\selectfont\bfseries}}
\renewcommand{\@seccntformat}[1]{\csname the#1\endcsname.\nobreakspace}
\makeatother

\usepackage{colortbl}

\definecolor{linkblue}{rgb}{0.21,0.49,0.74}
\definecolor{firstc}{RGB}{214,230,243}
\definecolor{secondc}{RGB}{252,238,190}
\newcommand{\first}[1]{\cellcolor{firstc}#1}
\newcommand{\second}[1]{\cellcolor{secondc}#1}
\definecolor{deltaup}{rgb}{0.10,0.55,0.25}
\definecolor{deltadown}{rgb}{0.80,0.15,0.15}
\usepackage[pagebackref,breaklinks,colorlinks,allcolors=linkblue]{hyperref}
\usepackage[capitalize]{cleveref}
\crefname{section}{Sec.}{Secs.}
\Crefname{section}{Section}{Sections}
\crefname{table}{Tab.}{Tabs.}
\Crefname{table}{Table}{Tables}
\crefname{figure}{Fig.}{Figs.}
\Crefname{figure}{Fig.}{Figs.}
\hypersetup{
  pdftitle={SciWorldBench: Do Video World Models Really Understand Scientific Experiments?},
  pdfauthor={Xue Yang, Haonan Cai, Xinyu Hang, Zhimeng Huang, Xuanyi Liu, Chuanmin Jia, Siwei Ma}
}

\title{SciWorldBench: Do Video World Models Really Understand\\Scientific Experiments?}

\newcommand{\authorprofile}[2]{%
  \href{https://openreview.net/profile?id=~#1}{\textcolor{black}{#2}}%
  \hyperlink{affiliation-pku}{\textcolor{black}{\textsuperscript{1}}}}
\author{%
  \authorprofile{Xue_Yang19}{Xue Yang},
  \authorprofile{Haonan_Cai1}{Haonan Cai},
  \authorprofile{Xinyu_Hang1}{Xinyu Hang}\footnotemark[1],
  \authorprofile{Zhimeng_Huang1}{Zhimeng Huang}\footnotemark[1],
  \authorprofile{Xuanyi_Liu3}{Xuanyi Liu},
  \authorprofile{Chuanmin_Jia1}{Chuanmin Jia},
  \authorprofile{Siwei_Ma1}{Siwei Ma}\\[3pt]
  {\normalsize\hypertarget{affiliation-pku}{\textsuperscript{1}\,Peking University}}%
}

\begin{document}
\twocolumn[{%
\renewcommand\twocolumn[1][]{#1}%
\maketitle
\begin{center}
  \includegraphics[width=\textwidth]{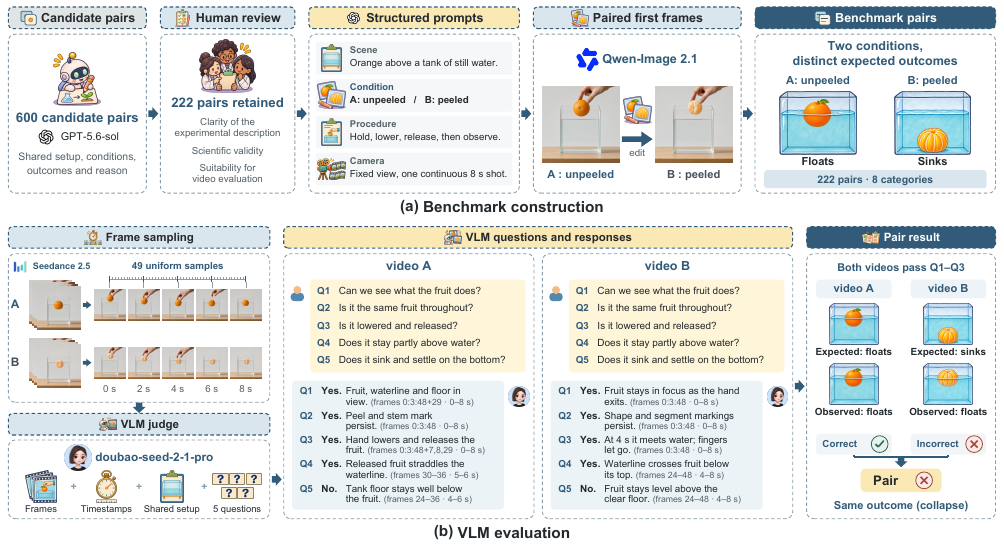}
  \captionof{figure}{Overview of SciWorldBench. Top: benchmark construction. Bottom:
VLM evaluation. The prompts, five VLM scoring questions, and answers shown in the figure are
shortened for illustration.}
  \label{fig:overview}
\end{center}
\vspace{0.5em}
}]
\footnotetext[1]{Work done while visiting Nanyang Technological University, Singapore.}
\begin{abstract}
While the ability of video generation models to accurately depict scientific
phenomena has received growing attention, determining whether these models
genuinely understand the underlying principles remains a
critical challenge. Existing benchmarks typically evaluate video samples in
isolation, making it difficult to expose models that simply fit superficial
visual priors. In scientific experiments, a small change in one condition
produces a markedly different outcome even when the rest of the setup remains
fixed. Capitalizing on this sensitivity, we introduce
SciWorldBench, a benchmark comprising paired scientific experiments. Each pair
varies a single condition in an otherwise fixed setup, yielding two mutually
exclusive expected outcomes. Each generation prompt specifies the condition
but omits the expected outcome. We evaluate
ten text-and-image-to-video (TI2V) models using an independent vision-language
model (VLM) that assesses each video without access to the condition text or
the label identifying the correct outcome. A pair is scored as correct only
when both videos depict their respective scientifically correct outcomes.
Across all ten models, we find that the two videos in an experimental pair
frequently depict the same outcome even though the conditions require different
results. This recurring failure suggests that current video world models
struggle to apply the underlying scientific principles reliably during
generation.
\end{abstract}

\clearpage
\begin{figure*}[t]
  \centering
  \scriptsize
  \colorbox{firstc}{\parbox{\dimexpr\textwidth-2\fboxsep}{\footnotesize\textbf{Same first frame: the condition appears only in the prompt.}}}\\[3pt]
    \begin{minipage}[t]{0.24\textwidth}\centering
      \textbf{Ice cube released into a liquid}\\{\color{gray}Fluids}\\[2pt]
      \includegraphics[width=0.46\linewidth]{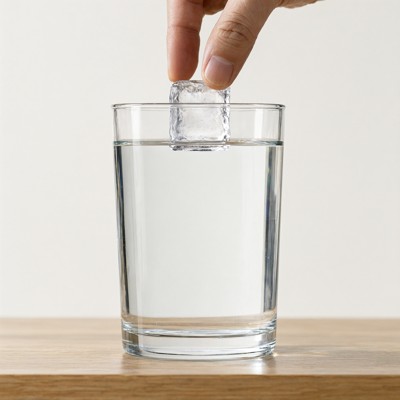}\hspace{2pt}%
      \includegraphics[width=0.46\linewidth]{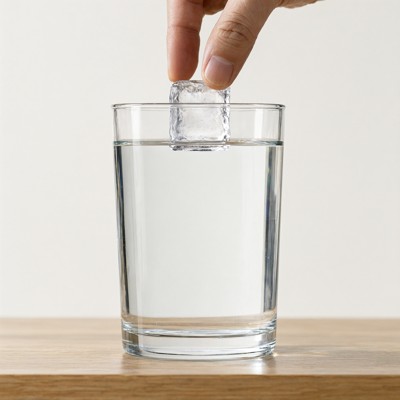}\\[-1pt]
      \parbox[t]{0.46\linewidth}{\centering{\color{gray}first frame A}\\\textbf{A}: water}\hspace{2pt}%
      \parbox[t]{0.46\linewidth}{\centering{\color{gray}first frame B}\\\textbf{B}: alcohol}
    \end{minipage}
\hfill
    \begin{minipage}[t]{0.24\textwidth}\centering
      \textbf{Glass pane struck once}\\{\color{gray}Materials}\\[2pt]
      \includegraphics[width=0.46\linewidth]{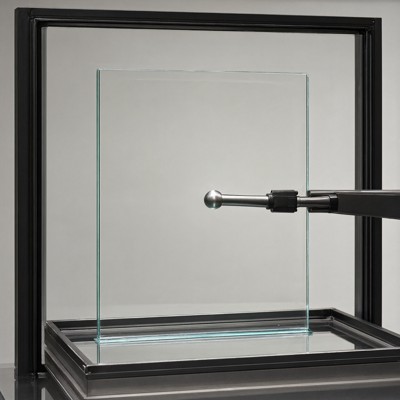}\hspace{2pt}%
      \includegraphics[width=0.46\linewidth]{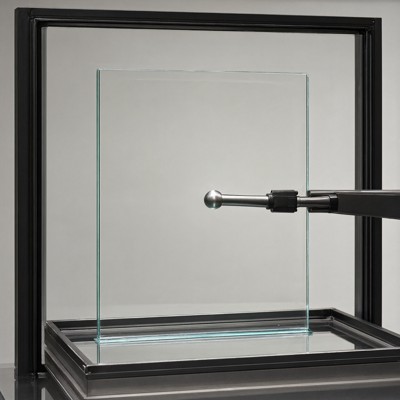}\\[-1pt]
      \parbox[t]{0.46\linewidth}{\centering{\color{gray}first frame A}\\\textbf{A}: annealed glass}\hspace{2pt}%
      \parbox[t]{0.46\linewidth}{\centering{\color{gray}first frame B}\\\textbf{B}: tempered glass}
    \end{minipage}
\hfill
    \begin{minipage}[t]{0.24\textwidth}\centering
      \textbf{Magnet released on a door}\\{\color{gray}Electromagnetism}\\[2pt]
      \includegraphics[width=0.46\linewidth]{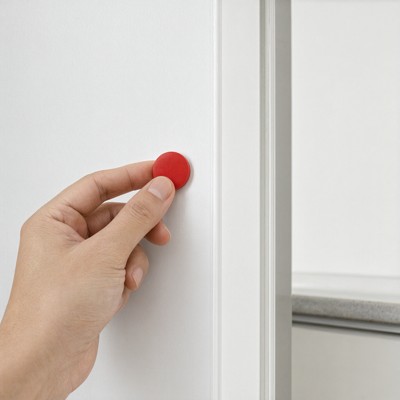}\hspace{2pt}%
      \includegraphics[width=0.46\linewidth]{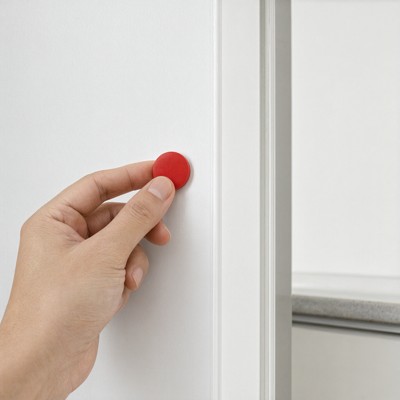}\\[-1pt]
      \parbox[t]{0.46\linewidth}{\centering{\color{gray}first frame A}\\\textbf{A}: ordinary steel}\hspace{2pt}%
      \parbox[t]{0.46\linewidth}{\centering{\color{gray}first frame B}\\\textbf{B}: nonmagnetic stainless steel}
    \end{minipage}
\hfill
    \begin{minipage}[t]{0.24\textwidth}\centering
      \textbf{Glass rod placed in a liquid}\\{\color{gray}Optics}\\[2pt]
      \includegraphics[width=0.46\linewidth]{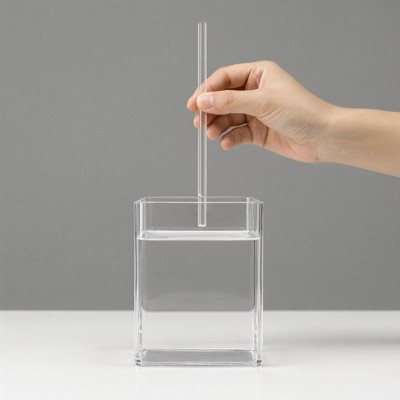}\hspace{2pt}%
      \includegraphics[width=0.46\linewidth]{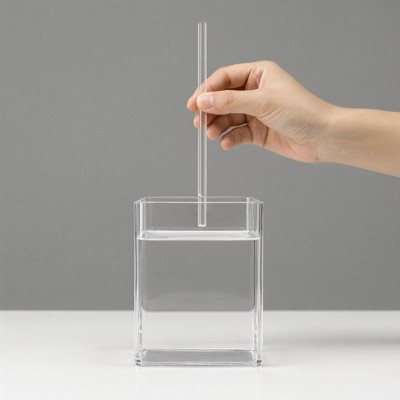}\\[-1pt]
      \parbox[t]{0.46\linewidth}{\centering{\color{gray}first frame A}\\\textbf{A}: water}\hspace{2pt}%
      \parbox[t]{0.46\linewidth}{\centering{\color{gray}first frame B}\\\textbf{B}: index-matched liquid}
    \end{minipage}
  \\[6pt]
  \colorbox{firstc}{\parbox{\dimexpr\textwidth-2\fboxsep}{\footnotesize\textbf{Different first frames: the condition is visible.}}}\\[3pt]
    \begin{minipage}[t]{0.24\textwidth}\centering
      \textbf{Roller on rising, diverging rails}\\{\color{gray}Mechanics}\\[2pt]
      \includegraphics[width=0.46\linewidth]{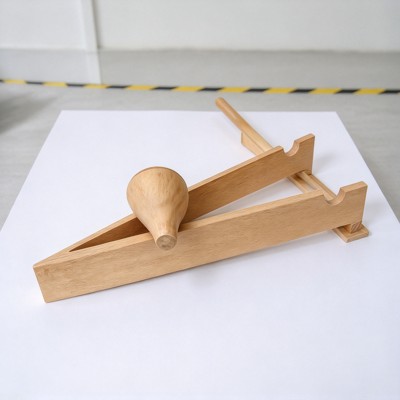}\hspace{2pt}%
      \includegraphics[width=0.46\linewidth]{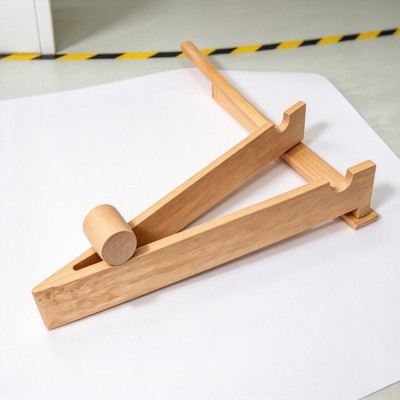}\\[-1pt]
      \parbox[t]{0.46\linewidth}{\centering{\color{gray}first frame A}\\\textbf{A}: double cone}\hspace{2pt}%
      \parbox[t]{0.46\linewidth}{\centering{\color{gray}first frame B}\\\textbf{B}: cylinder}
    \end{minipage}
\hfill
    \begin{minipage}[t]{0.24\textwidth}\centering
      \textbf{Cube dropped into warm water}\\{\color{gray}Thermodynamics}\\[2pt]
      \includegraphics[width=0.46\linewidth]{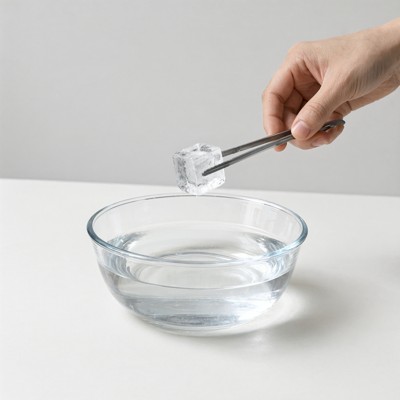}\hspace{2pt}%
      \includegraphics[width=0.46\linewidth]{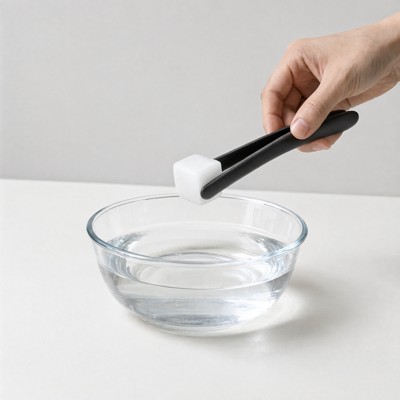}\\[-1pt]
      \parbox[t]{0.46\linewidth}{\centering{\color{gray}first frame A}\\\textbf{A}: water ice}\hspace{2pt}%
      \parbox[t]{0.46\linewidth}{\centering{\color{gray}first frame B}\\\textbf{B}: dry ice}
    \end{minipage}
\hfill
    \begin{minipage}[t]{0.24\textwidth}\centering
      \textbf{Water poured on a sponge}\\{\color{gray}Chemistry}\\[2pt]
      \includegraphics[width=0.46\linewidth]{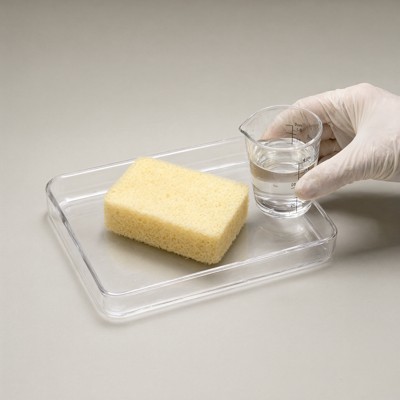}\hspace{2pt}%
      \includegraphics[width=0.46\linewidth]{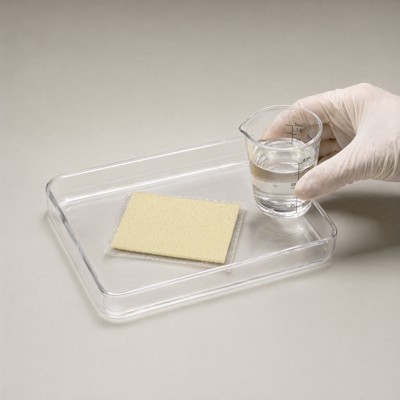}\\[-1pt]
      \parbox[t]{0.46\linewidth}{\centering{\color{gray}first frame A}\\\textbf{A}: ordinary sponge}\hspace{2pt}%
      \parbox[t]{0.46\linewidth}{\centering{\color{gray}first frame B}\\\textbf{B}: compressed sheet}
    \end{minipage}
\hfill
    \begin{minipage}[t]{0.24\textwidth}\centering
      \textbf{Water on a leaf}\\{\color{gray}Biology}\\[2pt]
      \includegraphics[width=0.46\linewidth]{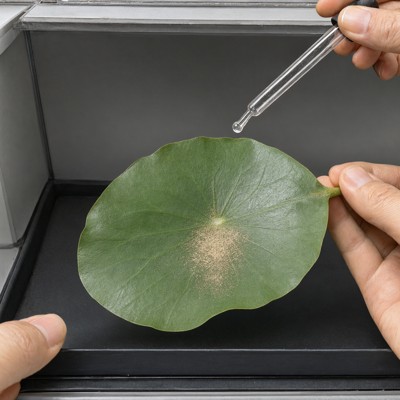}\hspace{2pt}%
      \includegraphics[width=0.46\linewidth]{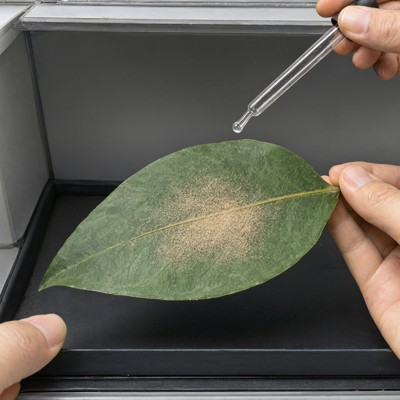}\\[-1pt]
      \parbox[t]{0.46\linewidth}{\centering{\color{gray}first frame A}\\\textbf{A}: lotus leaf}\hspace{2pt}%
      \parbox[t]{0.46\linewidth}{\centering{\color{gray}first frame B}\\\textbf{B}: ordinary leaf}
    \end{minipage}
  \caption{Example pairs from SciWorldBench. Each pair varies one condition while
holding the setup fixed; the conditions are listed below the images. Top:
both sides use the same first frame, and the condition is specified only in
the prompt. Bottom: the condition is visible in the first frame.}
  \label{fig:example}
\end{figure*}

\section{Introduction}
\label{sec:intro}

Text-and-image-to-video (TI2V) generation has advanced rapidly~\cite{veo,seedance,kling,wan3,vidu,minimaxh3,cosmos,ltx2,wan}.
Recent models, including Seedance 2.5~\cite{seedance}, Kling 3.0~\cite{kling},
Wan 3.0~\cite{wan3}, and Vidu Q3 Pro~\cite{vidu}, generate high-resolution videos
with realistic appearance and smooth motion from a first frame and a text
prompt. These models are increasingly used as video world models to simulate
scene evolution. In scientific experiments, changing a single condition can lead to a
markedly different outcome while the rest of the setup remains unchanged. Consider an orange released into
water: air in the intact peel lowers the fruit's average density, allowing it
to float, whereas a peeled orange sinks. The two experiments otherwise share
the same setup. In the example shown in \cref{fig:overview}, Seedance 2.5 generates videos
in which the orange floats under both conditions, although the peeled orange
should sink. Generating the correct outcome under one condition therefore does not
establish that the model understands how the experimental condition affects
the outcome. This failure motivates our central question: do video world
models really understand scientific experiments?

Existing benchmarks broaden task coverage and refine evaluation
criteria~\cite{videophy,phyworldbench} to assess whether models can generate
physically correct videos from prompts specifying scenes, events, and
outcomes~\cite{videophy2,t2vphysbench}. However, generating the correct outcome
does not establish that the model derived it from the given conditions
through causal reasoning. A complementary line of work examines responses
to controlled changes in autonomous driving and robotic
manipulation~\cite{whatifworld} and simulated physical
scenes~\cite{cronos,geoflip,causalcascade}. Their focus on embodied interactions
and mechanical events provides limited evidence of whether models understand
condition--outcome relationships across a broader range of scientific
experiments.

We introduce SciWorldBench, a benchmark of 222 experiment pairs across
eight categories. Each pair holds the setup fixed and varies one condition,
with mutually exclusive expected outcomes. Generation prompts specify the
condition but omit the outcome. Half of the pairs use an identical first
frame, so the condition is specified only in text. In the other half, the
condition is visible in the first frame. The two videos are generated
in separate requests, and a pair passes only when both show their respective
scientifically correct outcomes.
A VLM uses visual evidence to check experimental prerequisites and
separately assess whether each candidate outcome occurs. We score a video
as correct only when all prerequisite checks pass and the judgments confirm
the expected outcome while ruling out the alternative.

Across ten TI2V models, Paired Accuracy (PA) ranges from 1.4\% to 24.8\%,
and the highest Scientific Outcome Accuracy (SOA) is 53.8\%.
A recurring failure is that models generate the same outcome despite the
contrasting conditions: among pairs in which both videos pass the prerequisite
checks and each shows exactly one outcome, the Collapse Rate (CR) is
55.7\%--91.7\%. Providing the correct outcome in the prompt
improves generation for all six tested open-weight models, but
most still tend to generate the same outcome under both conditions.
The highest PA is 33.3\%, and CR exceeds 50\% for five of the six models.
Both Cosmos models combine understanding and generation, yet their
ability to infer experimental outcomes in text does not reliably carry over
to video generation.
Cosmos3-Super, for example, predicts both outcomes correctly in text
for 63.1\% of pairs but generates both correctly in only 9.0\%.

Our contributions are as follows.
\begin{enumerate}
\item  
We introduce SciWorldBench, a benchmark of 222 experiment pairs across
eight categories for evaluating whether video world models understand
scientific experiments. Each pair specifies contrasting experimental
conditions and their distinct, visually verifiable outcomes.
\item 

We establish an evaluation protocol that distinguishes unmet experimental
prerequisites from incorrect outcomes and assesses correctness under each
condition. The VLM identifies the outcome shown in a video without receiving
the condition text or the correct-outcome label, and we validate its
judgments against human consensus.
\item 

We evaluated 10 mainstream TI2V models on our benchmark. Across all the
models, the highest PA was 24.8\%, and paired videos frequently showed the
same outcome even though the two conditions required different results.
These findings show that current video world models lack reliable scientific
understanding in generation.
\end{enumerate}

\section{Related Work}
\label{sec:related}

\paragraph{Physical fidelity in video generation.}
General video generation benchmarks evaluate visual quality and text--video
alignment~\cite{vbench,vbenchpp,evalcrafter}, while other benchmarks examine
temporal change and compositionality~\cite{chronomagic,tcbench,t2vcompbench}.
VBench-2.0~\cite{vbench2} includes physics and common sense in its evaluation of
intrinsic faithfulness. Benchmarks dedicated to physical fidelity assess whether
generated actions and material interactions obey physical
laws~\cite{videophy,phygenbench,physvidbench,worldbench}.
Some tasks specify both the event and its outcome in the prompt, testing how
accurately a model depicts the prescribed motion~\cite{videophy2,t2vphysbench,phyworldbench}.
Other benchmarks evaluate physical prediction by comparing generated
continuations with recordings of real experiments~\cite{physicsiq,physicsiqverified}
or measuring whether generated dynamics satisfy conservation laws~\cite{morpheus}.
Physically correct motion alone, however, does not establish that the model
derived the outcome from the given conditions through causal reasoning.
When the prompt explicitly states the outcome, the model can depict it without
making that inference, leaving its understanding of the causal relationship
untested.

\paragraph{Scientific understanding and reasoning in video generation.}
Scientific reasoning has been studied both in the interpretation of recorded
experiments~\cite{expvid,scivideobench,beaker} and in the generation of
scientific phenomena~\cite{risevideo}.
VideoScience-Bench~\cite{videoscience} evaluates whether models generate the
phenomena predicted by interacting concepts in physics and chemistry.
Sci-VBench~\cite{scivbench} uses expert criteria to assess whether generated
processes are scientifically and causally correct across disciplines.
A correct result for one experimental setup does not establish whether the model
can also generate the appropriate alternative when one condition changes.

\paragraph{Condition dependence of generated outcomes.}
Prior work examines how changes in physical parameters affect generated
motion~\cite{physparamcontrol} and whether a single set of physical parameters
can explain the trajectories predicted under different actions from the same
initial state~\cite{oneworld}.
CRONOS~\cite{cronos} studies physical consistency as viewpoint or appearance
varies while the event type remains fixed.
Other work examines how interventions alter driving and robotic manipulation
outcomes~\cite{whatifworld}, how initial geometry determines mechanical
outcomes~\cite{geoflip}, and how changes propagate through a sequence of
physical interactions~\cite{causalcascade}.
These benchmarks focus primarily on embodied interactions and mechanical events,
leaving open whether models understand how conditions determine outcomes across
a broader range of scientific experiments.
SciWorldBench examines this question across eight experimental categories by
pairing two conditions of the same experiment that require mutually exclusive
outcomes.

\section{SciWorldBench}
\label{sec:benchmark}

SciWorldBench tests whether video world models understand scientific
experiments by evaluating whether they generate the scientifically correct
outcome under each condition. The benchmark contains 222 pairs of experiments. Within each pair, the
setup is fixed except for one condition, and the two conditions have different
expected outcomes. We present the task formulation in
\cref{sec:task}, benchmark construction in \cref{sec:construction}, the
evaluation protocol in \cref{sec:protocol}, and the metrics in
\cref{sec:metrics}.

\begin{figure*}[t]
  \centering
  \includegraphics[width=\textwidth]{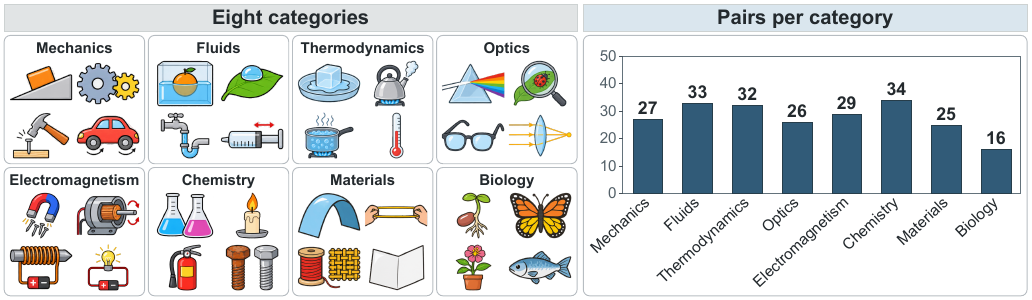}
  \caption[Categories of SciWorldBench.]{Categories of SciWorldBench.
    The left panel illustrates the eight categories with
    representative objects and phenomena.
    The bars on the right show how the 222 experimental pairs are distributed
    across them.}
  \label{fig:categories}
\end{figure*}

\subsection{Task Formulation}
\label{sec:task}

We represent a pair as $p = (S, c_A, c_B, I_A, I_B, o_A, o_B)$. The shared setup $S$ contains the scene,
procedure, and camera requirements shared by the two sides. Conditions $c_A$ and $c_B$ distinguish the
two sides and are expressed as short clauses in the prompts. A condition may
specify a property, such as a material, temperature, or liquid, or an action,
such as whether a bottle is squeezed. $I_A$ and $I_B$ denote the first frames
of the two sides. $o_A$ and $o_B$ denote the corresponding expected outcomes,
which are mutually exclusive and visually assessable from the frames.

For side $k \in \{A, B\}$, a TI2V model $G$ receives the first frame $I_k$ and
the prompt $[S; c_k]$ and generates a video $v_k = G(I_k, [S; c_k])$. The prompt specifies the condition but omits the outcome, requiring the
model to generate the result without explicit outcome guidance. The two
videos are generated in separate requests. Each includes only the first frame and
prompt for its own side and provides no access to the other condition. A video $v_k$ is correct if it
passes all three prerequisite checks, shows $o_k$, and
does not show the other outcome. A pair passes only when both $v_A$ and $v_B$
are correct.

\subsection{Benchmark Construction}
\label{sec:construction}

We use GPT-5.6-sol~\cite{gpt56} to generate 600 candidate pairs of 8-second
video experiments, following the pipeline in \cref{fig:overview}. Each
candidate defines a shared experimental setup and two alternative conditions,
which are combined to form a pair of video generation prompts. Each candidate
also specifies the expected outcome under each condition and a brief
explanation of the causal relationship between the condition and its expected
outcome. We then conduct expert review to assess the candidates along three
dimensions. (1) For \emph{clarity of the experimental description}, reviewers
check that each candidate specifies the shared setup and how the two conditions
differ. (2) For \emph{scientific validity}, they verify that each condition
leads to its stated outcome under that setup. (3) To assess
\emph{suitability for video evaluation}, they check whether each outcome is
observable within eight seconds and distinguishable from the alternative using
only the video frames. After review and refinement, 222 pairs
are retained. We then use GPT-6 Astra~\cite{gpt6} to reformulate the video
generation prompts into a structured format with four labeled fields: scene,
condition, procedure, and camera requirements. The two video generation
prompts in each pair differ only in the condition field, and neither
explicitly specifies the expected outcome or its explanation.

In the TI2V setting, each experimental condition is associated with a first
frame. We generate and edit these reference frames using Qwen-Image
2.1~\cite{qwenimage}. The side-A frame is generated from the shared
experimental setup and its corresponding condition. When the condition is
visually distinguishable, the side-B frame is edited from the side-A frame
while preserving the scene layout, camera viewpoint, and unrelated objects.
When the condition is not visually distinguishable in a static image, as in
the ice cube experiments with water and alcohol (\cref{fig:example}), the same
first frame is used for both sides, with the difference specified only in the
generation prompts. All reference frames are initially created at
$2048 \times 2048$ pixels and resized to match each model's output resolution
before video generation.

Finally, we organize the 222 pairs into eight categories based on the
underlying physical mechanisms (\cref{fig:categories}). We also divide them
into two equally sized subsets according to whether the paired videos share
the same first frame: 111 same-first-frame pairs and 111 different-first-frame
pairs.

\subsection{Evaluation Protocol}
\label{sec:protocol}

We use doubao-seed-2-1-pro~\cite{seed2} at temperature 0 to evaluate the
two videos in each experimental pair separately (\cref{fig:overview}). For
each video, the VLM receives 49 full-resolution frames sampled uniformly
over time, their timestamps, and a description of the shared experimental
setup. We withhold the condition-specific text, the name of the generating
model, the A/B side label, and the label identifying the expected outcome.

We use the same five questions to score both videos in a pair. We shuffle the question order
for each video using a fixed seed. Three questions check whether the
target's behavior is visible, whether it remains the same object throughout
the video, and whether the experimental preconditions are met. The other two
questions each ask whether the frames show a particular candidate outcome.
We describe each outcome in terms of observable behavior, removing
condition-specific attributes and referring to the target in neutral terms.
For example, we rewrite ``The peeled orange sinks to the bottom'' as ``The round
fruit sinks through the water and settles on the bottom of the tank.''

For each question, the VLM returns an answer, a brief observation, and a
rationale in a fixed JSON format.
Both \emph{yes} and \emph{no} require the
indices of at least two supporting frames and the time interval they span.
When the visual evidence is insufficient, the VLM answers \emph{unknown}.
We score a video as correct only if all three prerequisite checks pass and
the VLM answers \emph{yes} for the outcome expected under that condition
and \emph{no} for the alternative.

\subsection{Metrics}
\label{sec:metrics}

\begin{table*}[!t]
  \centering
  \papertablestyle
  \begin{tabularx}{\linewidth}{@{}lllYYYY@{}}
    \toprule
    Model & Type & Params & PPR $\uparrow$ & SOA $\uparrow$ & PA $\uparrow$ & CR $\downarrow$ \\
    \midrule
    Seedance 2.5 & Commercial & -- & \first{94.6} & \first{53.8} & \first{24.8} & \first{55.7} \\
    Kling 3.0 & Commercial & -- & \second{92.6} & \second{45.0} & \second{15.8} & \second{66.9} \\
    Vidu Q3 Pro & Commercial & -- & 88.7 & 44.6 & 15.3 & 72.2 \\
    Wan 3.0 & Commercial & -- & 92.1 & 43.2 & 12.6 & 74.0 \\
    Cosmos3-Super & Open-weight & 64B & 90.1 & 37.8 & 9.0 & 79.3 \\
    MiniMax-H3 & Open-weight & 33B & 91.7 & 39.9 & 7.2 & 84.3 \\
    LTX-2.3 & Open-weight & 22B & 78.4 & 32.0 & 7.2 & 75.6 \\
    Cosmos3-Nano & Open-weight & 16B & 77.0 & 30.4 & 5.4 & 82.1 \\
    Wan2.2-I2V-A14B & Open-weight & 14B active & 75.0 & 28.8 & 3.6 & 87.1 \\
    Wan2.2-TI2V-5B & Open-weight & 5B & 48.0 & 16.2 & 1.4 & 91.7 \\
    \bottomrule
  \end{tabularx}
  \caption{Results on SciWorldBench (\%). PPR and SOA are computed over the
444 videos of each model, and PA over the 222 pairs. CR is computed over the
decided pairs, in which both videos pass the prerequisite checks and each
shows exactly one outcome.
We assess model performance across four complementary metrics
and rank models separately on each.
In each column, the best and second-best results are shaded light blue and
light yellow.}
  \label{tab:main}
  \par\smallskip
  \begin{minipage}{\linewidth}
    \captionsetup{type=figure}
    \centering
    \includegraphics{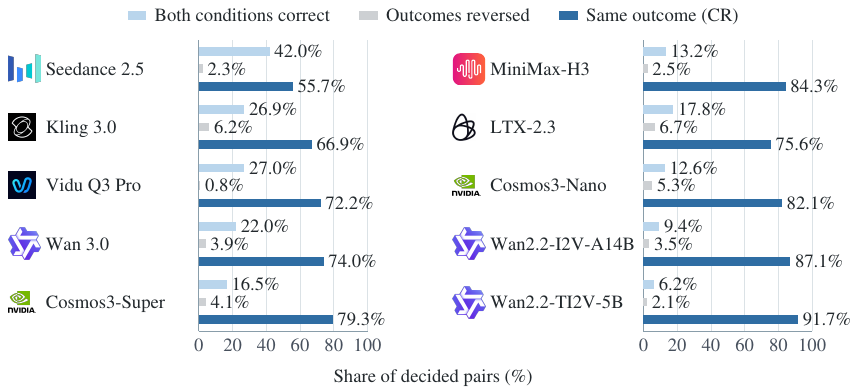}
    \caption{Outcome distribution among decided pairs for each model.
A pair is decided when both videos pass all prerequisite checks and each shows
exactly one candidate outcome. The bars show the percentages of these pairs
with both outcomes correct (light blue), the two expected outcomes reversed
(light gray), or the same outcome under both conditions (dark blue).
The last percentage is the Collapse Rate (CR).}
    \label{fig:radar}
  \end{minipage}
\end{table*}

We report four metrics for each model. Let $\mathcal{P}$ be the set of $n=222$
pairs. For a pair $p \in \mathcal{P}$, $v_A$ and $v_B$ are the videos generated for
its two sides. For a video $v$, let $g(v)=1$ if visibility, identity, and
execution are all answered yes and $g(v)=0$ otherwise, and let $q_A(v),
q_B(v) \in \{\text{yes}, \text{no}, \text{unknown}\}$ be the answers to the questions
on whether $v$ shows $o_A$ and $o_B$.

\paragraph{Prerequisite Pass Rate (PPR)} is the share of the $2n$ videos that pass all
three prerequisite checks:
\begin{equation}
\text{PPR} = \frac{1}{2n}\sum_{p \in \mathcal{P}} \sum_{k \in \{A,B\}} g(v_k).
\end{equation}
This metric reports the proportion of videos that satisfy the prerequisites
for outcome evaluation.

\paragraph{Scientific Outcome Accuracy (SOA)} is the share of the $2n$ videos that are correct. For
side $k \in \{A,B\}$, with $\bar{k}$ the other side, the
video $v_k$ is correct ($s_k=1$) when it passes the prerequisite checks, shows
its own outcome, and does not show the other one:
\begin{equation}
s_k = \begin{cases} 1, & \text{if } g(v_k)=1,\ q_k(v_k)=\text{yes, and } q_{\bar{k}}(v_k)=\text{no},\\ 0, & \text{otherwise.} \end{cases}
\end{equation}
SOA is
\begin{equation}
\text{SOA} = \frac{1}{2n}\sum_{p \in \mathcal{P}} (s_A + s_B).
\end{equation}
A video is scored as incorrect if any required answer is unknown or if
both outcomes are present. SOA measures success for individual conditions,
as in evaluations that score each prompt separately.

\paragraph{Paired Accuracy (PA)} is our main metric:
\begin{equation}
\text{PA} = \frac{1}{n}\sum_{p \in \mathcal{P}} s_A\, s_B.
\end{equation}
Evaluating the two sides independently can overestimate performance
across conditions: a commonly generated outcome may be correct for only one
side. PA counts a pair only when both outcomes are correct, requiring
successful generation under both experimental conditions. A model that
generates the same outcome under both conditions receives a pair score of 0.

\paragraph{Collapse Rate (CR)} quantifies how often a model generates the same outcome under both
conditions. If any prerequisite check fails, $\ell(v)$ is undefined. For videos
that pass all three checks ($g(v)=1$), we use the following rule.
\begin{equation}
\ell(v) =
\begin{cases}
  A, & \text{if } q_A(v){=}\text{yes and } q_B(v){=}\text{no},\\
  B, & \text{if } q_B(v){=}\text{yes and } q_A(v){=}\text{no},\\
  \text{undefined}, & \text{otherwise.}
\end{cases}
\end{equation}
A pair is decided when both videos pass the prerequisite checks and
$\ell(v_A)$ and $\ell(v_B)$ are both defined, and we write
$\mathcal{D}$ for the decided pairs. The CR is
\begin{equation}
\text{CR} = \frac{|\{p \in \mathcal{D} : \ell(v_A) = \ell(v_B)\}|}{|\mathcal{D}|}.
\end{equation}
PA measures whether a model generates the correct outcomes under both
experimental conditions. CR characterizes a specific failure among decided
pairs: both generated videos satisfy the prerequisite checks and show the
same clearly identifiable outcome, so the observed outcome is unchanged
across the two conditions. A lower CR indicates a smaller proportion of
same-outcome failures among decided pairs. It does not necessarily imply
higher PA, since the two outcomes can also be reversed.

\section{Experiments}
\label{sec:experiments}

\subsection{Setup}
\label{sec:setup}

We evaluate ten TI2V models. Four are commercial: Seedance 2.5~\cite{seedance},
Kling 3.0~\cite{kling}, Wan 3.0~\cite{wan3}, and Vidu Q3 Pro~\cite{vidu}. Six are open-weight:
MiniMax-H3 (33B)~\cite{minimaxh3}, Cosmos3-Super (64B) and Cosmos3-Nano (16B)~\cite{cosmos},
LTX-2.3 (22B)~\cite{ltx2}, and Wan2.2-I2V-A14B (mixture of experts, 14B active) and
Wan2.2-TI2V-5B (5B)~\cite{wan}.
For each condition, all models use the same prompt and source
image (\cref{sec:construction}).
For each pair, we generate one video per condition in separate requests using the same random seed (42) to reduce sampling variability when comparing outcomes across conditions.
We set the target video duration to 8 seconds for all ten models.
To make full use of each model's generation capabilities, we follow its
official recommendations for both output resolution and frame rate.

\subsection{Main Results}
\label{sec:main}
\label{sec:collapse}

Across the ten TI2V models, PA ranges from 1.4\% to 24.8\%
(\Cref{tab:main}). The four commercial models occupy the top four positions,
with PA ranging from 12.6\% to 24.8\%, compared with 1.4\% to 9.0\% for the six
open-weight models. Seedance 2.5 ranks first, followed by Kling 3.0 and Vidu Q3
Pro.
The four commercial models, Cosmos3-Super, and MiniMax-H3 pass all
three prerequisite checks in 88.7\%--94.6\% of their videos, yet their SOA
ranges from only 37.8\% to 53.8\%. Even Seedance 2.5, which generates the
scientifically correct outcome in 53.8\% of individual videos, does so under
both conditions in only 24.8\% of all experimental pairs.

Among decided pairs, all ten models more often repeat the same outcome
than generate different outcomes across conditions (\Cref{fig:radar}).
CR ranges from 55.7\% for Seedance 2.5 to 91.7\% for Wan2.2-TI2V-5B.
When a model repeats an outcome, it is correct under one condition but fails
to produce the different outcome required by the other.

\begin{table}[p]
  \begin{minipage}[c][\textheight][s]{\columnwidth}
  \centering
  \papertablestyle
  \newcommand{\guidancedelta}[2]{\,\textcolor{#1}{(#2)}}
  \begin{tabularx}{\linewidth}{@{}lYYY@{}}
    \toprule
    Model & PA $\uparrow$ & SOA $\uparrow$ & CR $\downarrow$ \\
    \midrule
    MiniMax-H3 & 7.2 & 39.9 & 84.3 \\
    \quad + outcome stated & 33.3\guidancedelta{deltaup}{+26.1} & 59.5\guidancedelta{deltaup}{+19.6} & 40.2\guidancedelta{deltadown}{$-$44.1} \\
    \midrule
    Cosmos3-Super & 9.0 & 37.8 & 79.3 \\
    \quad + outcome stated & 27.5\guidancedelta{deltaup}{+18.5} & 56.5\guidancedelta{deltaup}{+18.7} & 50.4\guidancedelta{deltadown}{$-$28.9} \\
    \midrule
    Cosmos3-Nano & 5.4 & 30.4 & 82.1 \\
    \quad + outcome stated & 19.4\guidancedelta{deltaup}{+14.0} & 47.5\guidancedelta{deltaup}{+17.1} & 58.6\guidancedelta{deltadown}{$-$23.5} \\
    \midrule
    LTX-2.3 & 7.2 & 32.0 & 75.6 \\
    \quad + outcome stated & 19.4\guidancedelta{deltaup}{+12.2} & 45.9\guidancedelta{deltaup}{+13.9} & 57.5\guidancedelta{deltadown}{$-$18.1} \\
    \midrule
    Wan2.2-I2V-A14B & 3.6 & 28.8 & 87.1 \\
    \quad + outcome stated & 14.0\guidancedelta{deltaup}{+10.4} & 40.1\guidancedelta{deltaup}{+11.3} & 69.1\guidancedelta{deltadown}{$-$18.0} \\
    \midrule
    Wan2.2-TI2V-5B & 1.4 & 16.2 & 91.7 \\
    \quad + outcome stated & 3.6\guidancedelta{deltaup}{+2.20} & 20.7\guidancedelta{deltaup}{+4.50} & 81.1\guidancedelta{deltadown}{$-$10.6} \\
    \bottomrule
  \end{tabularx}
  \caption{Results (\%) without and with explicit outcome guidance. For each model,
the first row reports the original setting and the second includes the
outcome in the prompt. Colored values in parentheses give percentage-point changes from the first row
(green for increases and red for decreases). CR is computed over
decided pairs separately for each setting.}
  \label{tab:told}
  \par\vfill

  \captionsetup{type=table}
  \begin{tabularx}{\linewidth}{@{}lYYYY}
    \toprule
    & \multicolumn{2}{c}{PA $\uparrow$} & \multicolumn{2}{c}{CR $\downarrow$} \\
    \cmidrule(lr){2-3}\cmidrule(l){4-5}
    Model & different & same & different & same \\
    \midrule
    Seedance 2.5 & \first{30.6} & \first{18.9} & \first{43.8} & \first{67.2} \\
    Kling 3.0 & \second{16.2} & \second{15.3} & \second{65.5} & \second{68.1} \\
    Vidu Q3 Pro & \second{16.2} & 14.4 & 68.4 & 75.4 \\
    Wan 3.0 & 14.4 & 10.8 & 67.2 & 79.7 \\
    Cosmos3-Super & 9.0 & 9.0 & 75.5 & 82.4\\
    MiniMax-H3 & 8.1 & 6.3 & 79.6 & 88.1\\
    LTX-2.3 & 8.1 & 6.3 & 65.8 & 82.7\\
    Cosmos3-Nano & 8.1 & 2.7 & 71.4 & 90.6\\
    Wan2.2-I2V-A14B & 6.3 & 0.9 & 80.6 & 91.8\\
    Wan2.2-TI2V-5B & 2.7 & 0.0 & 83.3 & 100.0\\
    \bottomrule
  \end{tabularx}
  \caption{Results by first-frame group (\%), 111 pairs per group. In each
column, the best and second-best results are shaded light blue and light yellow.}
  \label{tab:firstframe}
  \par\vfill
  \captionsetup{type=figure}
  \includegraphics[width=\linewidth]{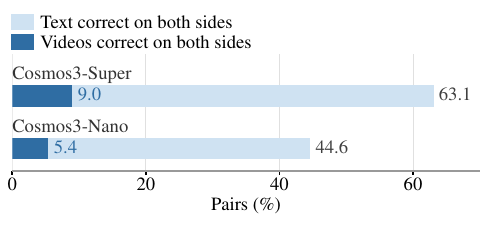}
  \caption{Textual outcome prediction and video generation for the same models
across 222 pairs. Light blue bars show the percentage of pairs with correct
text predictions under both conditions. Dark blue bars show video PA, the
percentage of pairs in which both videos depict their respective correct outcomes.}
  \label{fig:textvideo}
  \par\kern0pt
  \end{minipage}
\end{table}

\begin{table}[!t]
  \centering
  \papertablestyle
  \begin{tabularx}{\linewidth}{@{}lYYYY}
    \toprule
    Category & Pairs & SOA $\uparrow$ & PA $\uparrow$ & CR $\downarrow$ \\
    \midrule
    Mechanics & 27 & 30.2 & 8.1 & \second{71.3} \\
    Fluids & 33 & 37.4 & 10.6 & 73.9 \\
    Thermodynamics & 32 & 40.8 & 10.0 & 76.5 \\
    Optics & 26 & \first{45.6} & \first{18.1} & \first{65.8} \\
    Electromagnetism & 29 & 35.2 & 7.6 & 80.6 \\
    Chemistry & 34 & 32.4 & 6.2 & 83.6 \\
    Materials & 25 & 36.8 & 10.0 & 75.2 \\
    Biology & 16 & \second{42.2} & \second{14.4} & 72.6 \\
    \bottomrule
  \end{tabularx}
  \caption{Results by category (\%), with metrics pooled across all ten models.
The best and second-best results in each column are shaded light blue and
light yellow, respectively.}
  \label{tab:category}
\end{table}

\subsection{Outcome Stated in the Prompt}
\label{sec:told}

To test whether stating the expected outcome reduces this repetition,
we append the correct outcome to each condition's original prompt and regenerate
the videos with the six open-weight models. We score the new videos using the
same protocol.

With the outcome given, all six models achieve higher SOA and PA, and
all six show lower CR (\Cref{tab:told}). Averaged across models, PA rises from 5.6\% to 19.5\%.
MiniMax-H3 shows the largest improvement. Its PA rises from 7.2\% to 33.3\%,
the highest among the six models, and its CR falls from 84.3\% to 40.2\%.
Wan2.2-TI2V-5B shows the smallest reduction in CR, from 91.7\% to 81.1\%. Even
with explicit outcome guidance, five of the six models still generate the same
outcome under both conditions in more than half of their decided pairs.

\subsection{Predicting Experimental Outcomes in Text}
\label{sec:textpred}

We return to the original prompts, which omit the expected outcome,
and compare how often Cosmos3-Super and Cosmos3-Nano predict both outcomes
correctly in text with how often they depict both correctly in video.
For each condition, we provide the full original generation prompt, which
specifies the scene, condition, procedure, and camera requirements. We omit the first frame and ask what happens.
The question is identical for both conditions of a pair. We manually
assess each answer against the expected outcome for its condition.
Both models predict the two outcomes correctly in text far more often
than they generate them correctly in video (\Cref{fig:textvideo}).
Cosmos3-Super gives correct text answers for both conditions in 63.1\% of
pairs, compared with a video PA of 9.0\%. For Cosmos3-Nano, the corresponding
rates are 44.6\% and 5.4\%.
Even when these models correctly infer both outcomes in text,
they often fail to depict those outcomes in their respective videos.

\subsection{Results by First-Frame Group}
\label{sec:analysis}

\Cref{tab:firstframe} compares the two first-frame groups in the
original generation setting, with 111 experimental pairs in each group. When
the condition is visible in the first frame, PA ranges from 2.7\% to 30.6\%.
When the two conditions share a first frame and differ only in text, PA ranges
from 0.0\% to 18.9\%. Nine models have lower PA in the shared-frame group,
while Cosmos3-Super achieves 9.0\% in both groups. All ten models have higher
CR in the shared-frame group. The largest
difference occurs for Seedance 2.5, whose CR is 43.8\% with different first
frames and 67.2\% with a shared first frame.
This pattern suggests that models are less responsive to a change
in experimental conditions when the prompt provides the only cue to that change.

\subsection{Results by Scientific Category}
\label{sec:category_results}

Each of the eight scientific categories contains 16 to 34 experimental
pairs. We pool results across the ten models within each category
(\Cref{tab:category}). Optics has the highest PA (18.1\%) and the lowest CR
(65.8\%), while chemistry has the lowest PA (6.2\%) and the highest CR
(83.6\%). For individual videos, SOA ranges from 30.2\% in mechanics to 45.6\%
in optics. Even in optics, where models are most accurate on both individual
videos and pairs, nearly two thirds of decided pairs show the same outcome
under both conditions.

\section{Discussion}
\label{sec:discussion}

Our empirical findings expose a sharp divide between recognizing what should happen
and synthesizing how it unfolds. In this section, we trace this failure to two underlying
bottlenecks---outcome-repetition bias stemming from generation without reasoning,
and a modality disconnect between reasoning and generation.

\textbf{Outcome-repetition bias stems from generation without reasoning.}
Because this bias is systematic, persists under explicit instruction,
and occurs across independent generation requests, it reflects a shared prior
rather than cross-run copying. We attribute this bias to visual associations
that text-and-image-to-video (TI2V) models learn from observational videos.
During generation, the models rely more heavily on the association between a
scene's initial appearance and a familiar outcome than on the experimental
condition specified in the prompt. As a result, they favor the same outcome
in both videos of a pair even when the two conditions require different results.
Current models behave more like \emph{renderers} than world simulators,
producing visually realistic videos that often fail to show how changing an
experimental condition alters the outcome. Even when the expected outcome is explicitly stated in
the prompt, they merely translate that label into surface rendering, accounting
for their partial compliance without true reasoning.

\textbf{The text--video gap reflects a modality disconnect.} Both unified Cosmos
models predict the correct outcomes for the two experimental conditions far
more often in text than they depict them in video
(\cref{fig:textvideo,sec:textpred}). This gap may partly reflect an intrinsic
difference in task demands. In the text task, the model makes a low-dimensional
judgment about which outcome follows from the stated experimental condition.
Video generation poses a higher-dimensional reasoning problem because the model
must map that same condition to a visual trajectory that captures how the scene
evolves across space and time to produce the corresponding outcome. Even when
the prompt supplies the correct outcome, removing the need to infer it, the best
video PA among the six tested open-weight models is 33.3\% (\cref{tab:told}).
In the original setting, all ten models show higher CR for pairs distinguished
only by text than for pairs in which the condition difference is also visible
in the first frame (\cref{tab:firstframe}). Together, these findings suggest
that textual information sufficient to identify an outcome does not reliably
guide its realization in video. Whether this difficulty arises from a mismatch
between the representations used for inference and generation remains open.

\section{Conclusion}
\label{sec:conclusion}

We introduced SciWorldBench to test whether video world models understand
scientific experiments. In each of its 222 pairs, two versions of the same
experiment differ in one condition and have mutually exclusive expected outcomes.
A VLM assesses each video without access to the condition text or the
expected-outcome label. A pair passes only if both videos depict the
scientifically correct outcomes for their respective conditions. Across ten
TI2V models, paired accuracy (PA) ranges from 1.4\% to 24.8\%. Models often
repeat the same outcome under both conditions, particularly when the first
frame is shared and only the text distinguishes the two inputs. Stating the
correct outcome improves generation for all six tested open-weight models,
yet the highest PA is 33.3\%. The two Cosmos models predict both outcomes
correctly in text far more often than they depict them correctly in video.
Together, these results suggest that current video world models have a limited
understanding of how experimental conditions determine outcomes.

Future work could use paired training videos to help models learn how a change
in one experimental condition affects the outcome. When both videos start from
the same frame, the model must use the condition stated in the prompt to infer
which outcome should occur. A latent or symbolic representation of that
prediction could guide video synthesis and provide a reference for checking the
generated result. Beyond checking whether an individual video matches that
prediction and appears plausible, benchmarks should test whether each video in
a controlled pair depicts the scientifically correct outcome for its condition.

\paragraph{Limitations.} Two limitations warrant consideration. First, our
mechanistic interpretation is derived from behavioral outputs and text
predictions rather than from probing internal representations; the deeper
mechanisms through which video generation models produce these phenomena remain
to be investigated. Second, the accuracy of our evaluation depends heavily on
the VLM judge: even with a state-of-the-art model, its agreement with human
annotators remains imperfect, which can introduce scoring errors.

{
    \small
    \IfFileExists{references.tex}{
}{%
      \bibliographystyle{plain}
      \bibliography{main}}
}

\end{document}